\documentclass[conference]{IEEEtran}
\IEEEoverridecommandlockouts
\usepackage{cite}
\usepackage{amsmath,amssymb,amsfonts}
\usepackage{algorithmic}
\usepackage{graphicx}
\usepackage{textcomp}
\usepackage{xcolor}
\usepackage{algorithm}
\usepackage{caption}
\usepackage{subcaption}
\usepackage{url}
\def\BibTeX{{\rm B\kern-.05em{\sc i\kern-.025em b}\kern-.08em
    T\kern-.1667em\lower.7ex\hbox{E}\kern-.125emX}}
\begin{document}

\title{Balanced Split: A new train-test data splitting strategy for imbalanced datasets}
\author{\IEEEauthorblockN{Azal Ahmad Khan}
\IEEEauthorblockA{\textit{Department of Chemistry} \\
\textit{Indian Institute of Technology}\\
Guwahati, India \\
k.azal@iitg.ac.in}
}

\maketitle

\begin{abstract}
 Classification data sets with skewed class proportions are called imbalanced.  Class imbalance is a problem since most machine learning classification algorithms are built with an assumption of equal representation of all classes in the training dataset. Therefore to counter the class imbalance problem, many algorithm-level and data-level approaches have been developed. These mainly include ensemble learning and data augmentation techniques.
 
 This paper shows a new way to counter the class imbalance problem through a new data-splitting strategy called balanced split. Data splitting can play an important role in correctly classifying imbalanced datasets. We show that the commonly used data-splitting strategies have some disadvantages, and our proposed balanced split has solved those problems.
\end{abstract}


\section{Introduction}
The unequal representation of classes in the dataset is termed class imbalance. Class imbalance is a problem in machine learning that has been constantly worked on. There are constant studies conducted on class imbalamce problem \cite{guo2008class}, \cite{japkowicz2002class}, \cite{johnson2019survey}, \cite{kotsiantis2006handling}. This imbalance is problematic since most machine learning classification algorithms are built with an assumption of equal representation of all classes in the training dataset. Most of the works to counter the class imbalance problem can be classified into algorithm level or data level. 
Algorithm-level works include ensemble learning techniques \cite{dietterich2002ensemble}, \cite{polikar2012ensemble}. Like class imbalance in twitter dataset \cite{liu2017addressing}, margin theory \cite{feng2018class}, drug-target interaction \cite{ezzat2016drug}, bearing defect \cite{farajzadeh2016efficient}, bioinformatics \cite{yang2013sample}, fault diagnostic \cite{wu2018integrated}, spam detection in social networks \cite{zhao2020heterogeneous}, human activity recognition \cite{guo2021evolutionary}, rocket burst prediction \cite{yin2021strength}, software defect prediction \cite{sun2012using}, financial bankruptcy prediction \cite{faris2020improving}, web attacks \cite{zuech2021detecting}, consumer credit risk \cite{papouskova2019two}, etc are all examples of ensemble learning in imbalanced datasets. The use of ensemble learning in class imbalance is also being constantly surveyed and reviewed \cite{krawczyk2017ensemble}, \cite{galar2011review}, \cite{dong2020survey}.

Another way to counter class imbalance is by artificially generating samples of the minority class. This practice is becoming very common in the machine learning community. There has been development of many such techniques like SMOTE \cite{chawla2002smote}, SMOTE-ENN \cite{guan2021smote}, SMOTE-SVM \cite{wang2017novel}, Kmeans-SMOTE \cite{douzas2018improving}, Borderline-SMOTE \cite{han2005borderline}, ADASYN \cite{he2008adasyn}, etc. General adversarial networks(GANs)\cite{goodfellow2020generative}, and its versions like quantum GANs \cite{dallaire2018quantum}, SMOTified-GAN \cite{sharma2022smotified} have been a recent addition to these techniques. Apart from these, there is RUS which undersamples the majority class \cite{prusa2015using}.

But these approaches artificially generate data to counter class imbalance we present a technique to use original data in such a way that it removes the class imbalance. We present a data-splitting strategy that answers the two main problems in the existing splitting strategies. The two problems are: 
\begin{enumerate}
\item No guarantee of representation of minority class in the training dataset
\item Imbalance in the training dataset.
\end{enumerate}
Our proposed balanced split solves both of these problems.

Data splitting plays an important role in imbalanced datasets. In \cite{rajpurkar2017chexnet}, they randomly split the entire dataset into 80\% training and 20\% test. This led to questioning whether there are patient overlaps between the train, validation, and test sets. If yes, this could overestimate the model performance. Later they clarified there was no overlap.
These are very common mistakes but are usually ignored because of less development in the field.

Now we can see how important data splitting is compared to the amount of work done. There has not been any development in the splitting strategy as researchers have opted to artificially generate data. Our work can potentially pave the way for future work by attracting researchers to the topic.

The paper is structured as follows in section II, we present the working of balanced split, In section III, we show the results of the tests that evaluate random-split, stratified-split, and balanced-split on two different classification algorithms. Last, in section IV, we conclude and present future research ideas.

\section{Balanced Split}
In this section, we explain how our proposed balanced split works. The balanced split's main principle is that the training dataset should have equal samples from all classes. Here we present the steps of splitting:
\begin{enumerate}
\item Calculate the number of samples to be included in the training dataset(TrS) using training ratio(tr) and total samples(m).
\[TrS = tr \times m\]
\item Include an equal number of samples from each class. These should be equal to -
\[\frac{TrS}{N}\]
where N is the number of classes.
\end{enumerate}

But it must be noted that the training ratio has an upper limit while using balanced split. This is determined using the following formula:
\begin{equation}
\frac{m  \times tr}{N} < min(m(A), m(B).....m(N))
\end{equation}
Therefore we can say that the upper limit of the training ratio in balanced split depends on the minority class ratio and the number of classes. From the above equation, it can also be noted that the upper limit of the training ratio depends linearly on the minority class ratio. 

\begin{table*}[htbp]
\footnotesize
\centering
\caption{Parameter and values for the algorithms compared.}
\label{tab1}
\begin{tabular}{lll}
\hline
\textbf{Split strategy} & \textbf{Advantages} & \textbf{Disadvantages} \\
\hline
Random split &  & Neither guarantees the inclusion of minority class nor equal \\
 & & representation of all the classes in the training dataset \\
\hline
Stratified split & Guarantees inclusion of minority class in the training &  Does not include an equal number of samples from each \\
 & dataset & class in the training dataset\\
\hline
Balanced split & Guarantees the inclusion of minority class and equal & Has an upper limit to train ratio\\ 
        & representation of all classes in the training dataset & \\
\hline
\end{tabular}
\end{table*}

In table \ref{tab1}, we highlight the advantages and disadvantages of random, stratified, and balanced split. The only disadvantage of balanced split is that it has an upper limit to train ratio.

\begin{figure*}[h]
\caption{Working of different strategies in binary class data split.}
\centering
\includegraphics[width=0.9\textwidth]{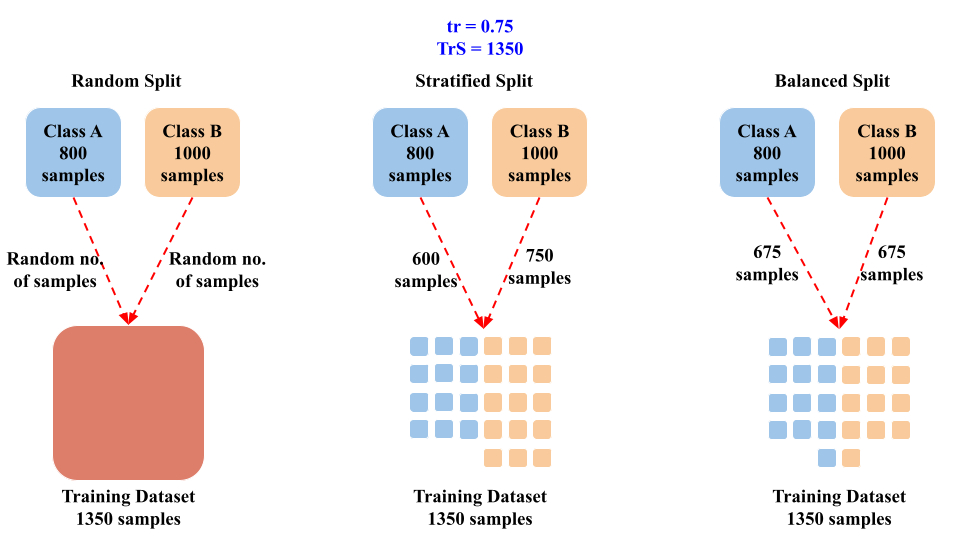}
\label{idiff}
\end{figure*}

Figure \ref{idiff} shows how different splitting strategies work. It is a binary classification example of class A, with 800 samples, and class B, with 1000 samples. The training ratio is set as 0.75, which means 75 percent of samples should be used in the training dataset and the remaining 25 percent for the test dataset. First, we show the case of a random split. Since it's the case of the random split, we can't determine the representation of each class in the training dataset. Next, comes stratified split here, we can determine the representation of each class in the training dataset. Stratified split guarantees the representation of each class in the training dataset, but it does not include an equal number of samples from each class in the training dataset except for the case of no imbalance. Next, we see the working of balanced split. Balanced split guarantees the inclusion of minority class and equal representation of all classes in the training dataset. The training ratio should be smaller than the upper limit, which can be determined by equation 1. For the example shown in figure \ref{idiff}, the upper limit of $tr$ is 0.88.

Algorithm \ref{alg} is the pseudo-code of balanced split. The pseudocode is for balanced split of datasets containing N classes and keeping the training ratio lesser than the upper limit. The code is published on GitHub and can be publicly accessed.

\begin{algorithm}
\caption{Pseudocode of Balanced Split}
\begin{algorithmic}
\label{alg}
\STATE Input: training ratio and the number of samples in the minority class.
\STATE Calculate the number of samples to be included from each class.
\FOR{$i = 1:N$}
\STATE Randomly select a specified number of samples from each class for the training dataset.
\STATE Include the remaining samples in the test dataset.
\ENDFOR
\STATE Output: balanced training and test dataset.
\end{algorithmic}
\end{algorithm}

\begin{figure*}
\centering
\begin{subfigure}{0.3\textwidth}
    \includegraphics[width=\textwidth]{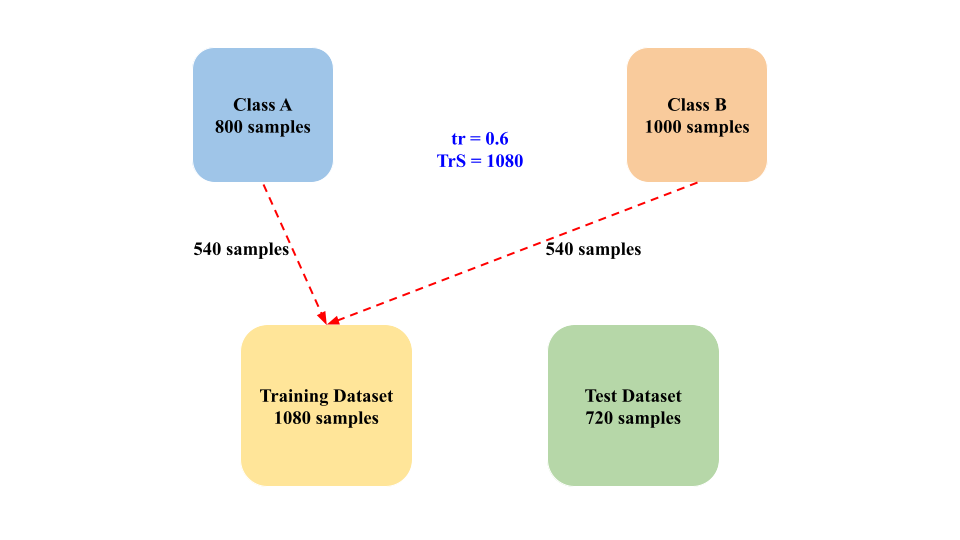}
    \caption{Balanced split with train ratio 0.6.}
    \label{i1}
\end{subfigure}
\hfill
\begin{subfigure}{0.3\textwidth}
    \includegraphics[width=\textwidth]{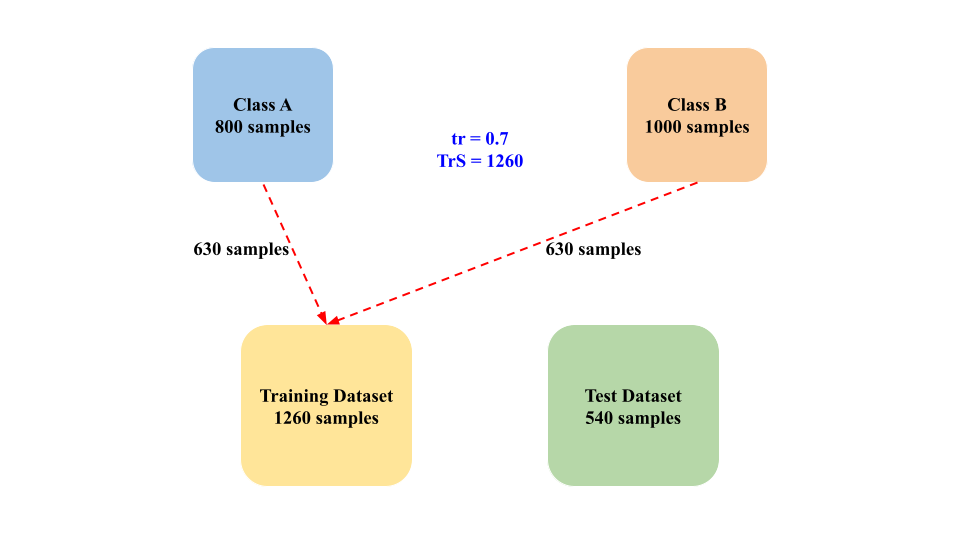}
    \caption{Balanced split with train ratio 0.7.}
    \label{i2}
\end{subfigure}
\hfill
\begin{subfigure}{0.3\textwidth}
    \includegraphics[width=\textwidth]{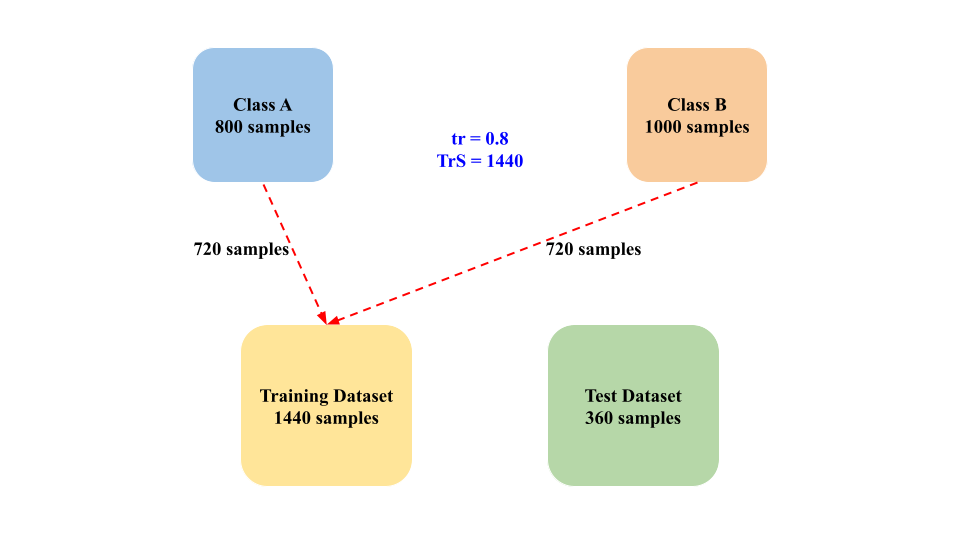}
    \caption{Balanced split with train ratio 0.8.}
    \label{i3}
\end{subfigure}
        
\caption{Working of balanced-split on a binary-class imbalanced dataset with different train ratios.}
\label{i123}
\end{figure*}

In figure \ref{i123}, we present the working of balanced split in the binary class dataset splitting. We take classes A and B in the example with 800 and 1000 samples, respectively. Class A is the minority class in the example as it has the least number of samples in the dataset. The minority class ratio is 0.44. Using equation 1, we can determine the upper limit of the training ratio, which is 0.89. In the figure, we show the split keeping three different train ratios. In figure \ref{i1}, tr was kept as 0.6, in figure \ref{i2}, tr was kept as 0.7, and in figure \ref{i3}, tr was kept as 0.8. All the training ratios were less than the upper limit, which is 0.89 hence balanced split was possible.

\begin{figure*}
\centering
\begin{subfigure}{0.3\textwidth}
    \includegraphics[width=\textwidth]{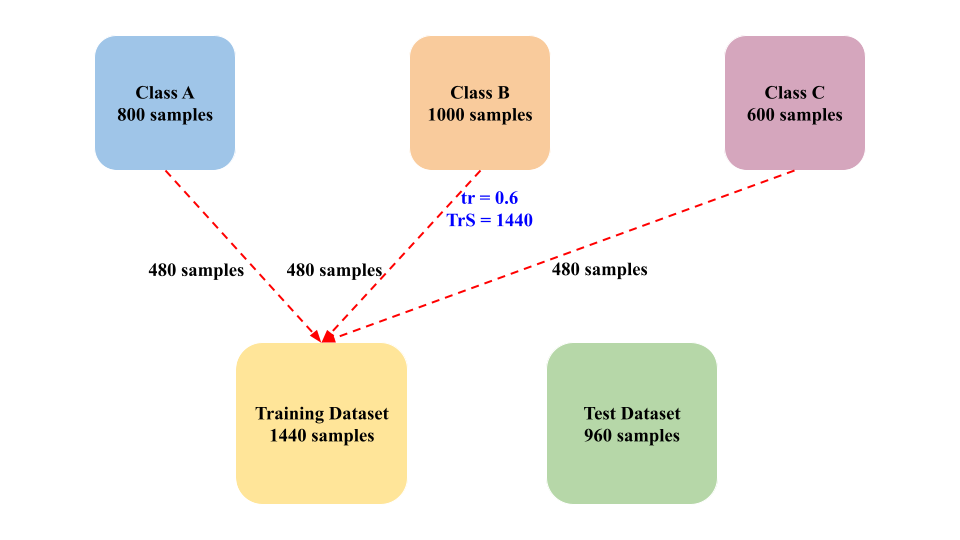}
    \caption{Balanced split with train ratio 0.6.}
    \label{i4}
\end{subfigure}
\hfill
\begin{subfigure}{0.3\textwidth}
    \includegraphics[width=\textwidth]{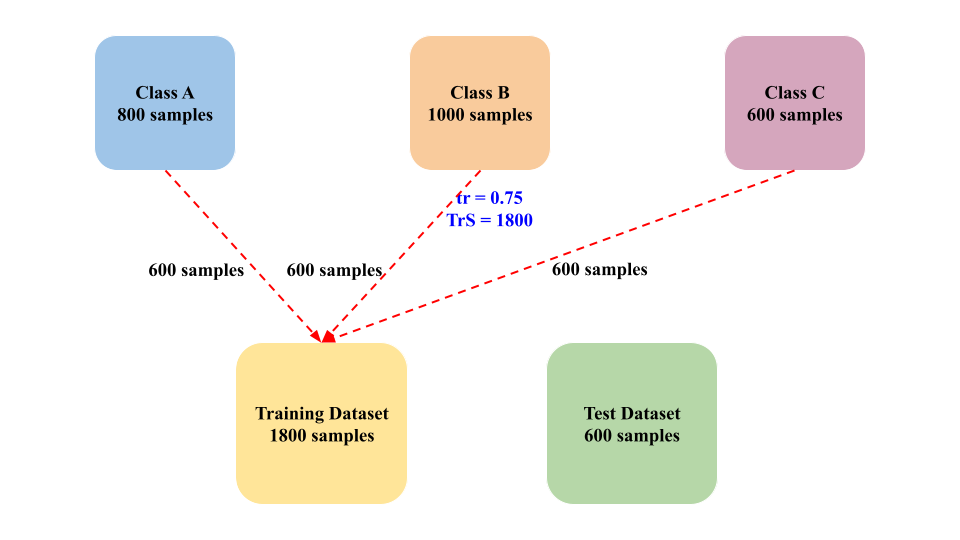}
    \caption{Balanced split with train ratio 0.75.}
    \label{i5}
\end{subfigure}
\hfill
\begin{subfigure}{0.3\textwidth}
    \includegraphics[width=\textwidth]{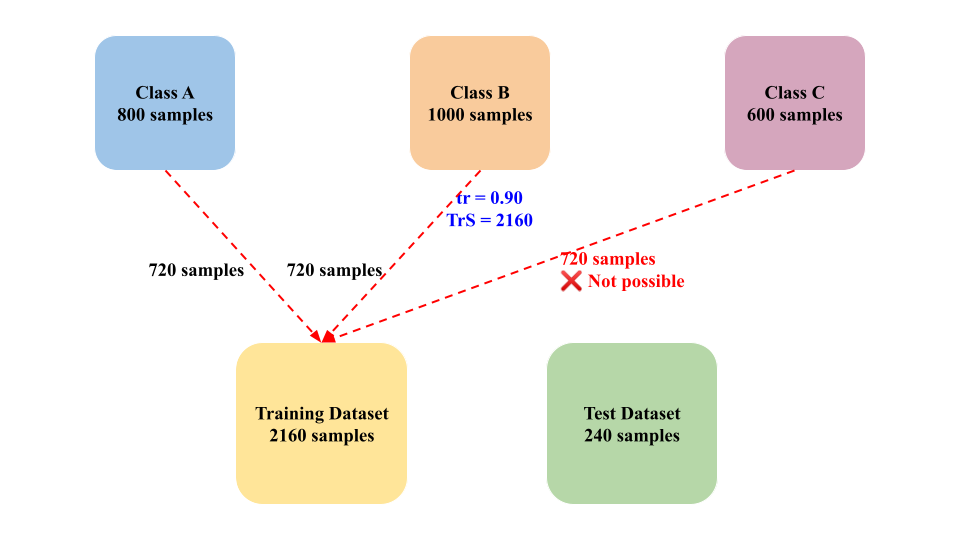}
    \caption{Balanced split with train ratio 0.9.}
    \label{i6}
\end{subfigure}
        
\caption{Working of balanced-split on a multi-class imbalanced dataset with different train ratios.}
\label{i456}
\end{figure*}

In figure \ref{i456}, we present the working of balanced split in the multi class dataset splitting. We take classes A, B, and C in the example with 800, 1000, and 600 samples, respectively. Class C is the minority class in the example as it has the least number of samples in the dataset. The minority class ratio is 0.25. Using equation 1, we can determine the upper limit of the training ratio, which is 0.75. In the figure, we show the split keeping three different train ratios. In figure \ref{i1}, tr was kept as 0.6, in figure \ref{i2}, tr was kept as 0.75, and in figure \ref{i3}, tr was kept as 0.9. 

In figure \ref{i5}, tr was kept as 0.75, equal to the upper limit. This leads to the inclusion of all the samples of minority class into the training dataset and leaving no samples for the test dataset. Therefore we would suggest users to refrain from using the upper limit as the training ratio to avoid exhausting all the minority samples in the training dataset. In figure \ref{i6}, tr was kept as 0.9, which is greater than the upper limit hence balanced split was not possible. 

\section{Tests and Results}
In this section, we show the results of the tests to evaluate balanced split. For comparison, we also evaluated random split and stratified split, which are the most common splitting strategies.

\subsection{Dataset description}
We conducted the tests on the customer dataset that is available on Kaggle. The dataset can be accessed using \url{https://www.kaggle.com/datasets/kaushiksuresh147/customer-segmentation?resource=download&select=Train.csv}. Dataset contains ten features and one target feature and 8058 total samples. The dataset is a multi-class imbalanced classification dataset. The dataset contains four classes, A, B, C, and D, with 1972, 1858, 1970, and 2268 samples. Class B is the minority class with a minority ratio of 0.23. Using equation 1, we got the upper limit of the training ratio, which is 0.92.
Gender, Ever\_Married, Graduated, Profession, Spending\_Score, and Var\_1 were categorical features that were converted to numerical values using a label encoder. Then, null values present in the features Work\_Experience and Family\_Size were replaced by the mode values of the particular features.

\subsection{Algorithms}
Two different algorithms were used in the classification. K-nearest neighbor was used. K-nearest neighbor is a supervised Learning technique. It's a non-parametric algorithm that does not takes assumptions. It classifies data into the specified number of neighbors by calculating their euclidean distances. Random forest classifier was used. It contains several decision trees on various subsets of the given dataset and takes the average to improve the predictive accuracy of that dataset. The hyperparameters were directly taken from scikit-learn library.

\subsection{Tests}
We have tested the splitting strategies with multiple train ratios. Nine test ratios were used from 0.5 to 0.9 with a step size of 0.05. All the train ratios are less than the upper limit, which is 0.92. We calculated the accuracy score and the weighted F-1 score. The F1-score combines the precision and recall of a classifier into a single metric by taking their harmonic mean. The F-1 score is the metric that is commonly used for imbalanced datasets.
\[F-1 = 2 \times \frac{precision \times recall}{precision + recall}\]

\begin{figure}
\centering
\begin{subfigure}{0.225\textwidth}
    \includegraphics[width=\textwidth]{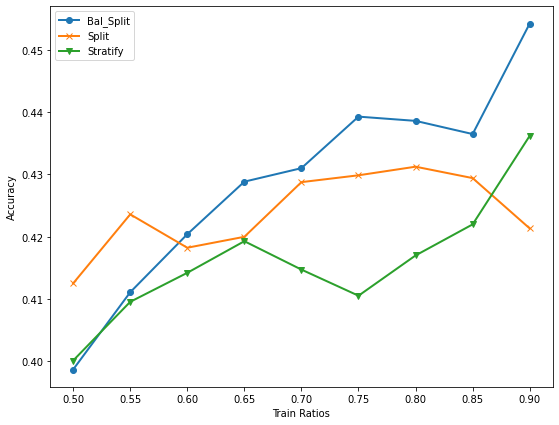}
    \caption{Comparison of accuracy of multiple splitting strategies.}
    \label{Knnacc}
\end{subfigure}
\hfill
\begin{subfigure}{0.225\textwidth}
    \includegraphics[width=\textwidth]{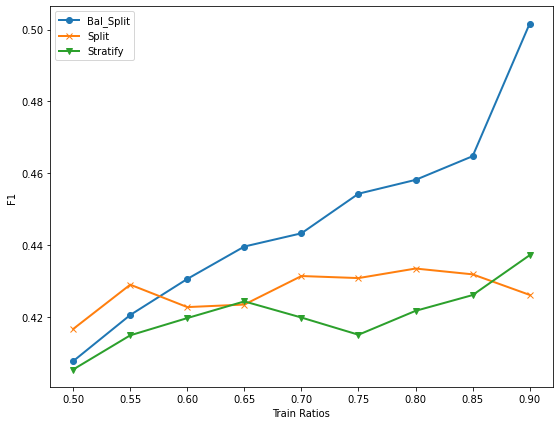}
    \caption{Comparison of F-1 score of multiple splitting strategies.}
    \label{Knnf1}
\end{subfigure}
\caption{Comparisons for K-nearest neighbor classifier.}
\label{Knn}
\end{figure}

The results of the K-nearest neighbor classifier's accuracy score and F-1 score are presented in table \ref{tab2}. Figure \ref{Knnacc} compares the random, stratified, and balanced split of the KNN classifier's accuracy score. Figure \ref{Knnf1} compares the random, stratified, and balanced split of the KNN classifier's F-1 score.

\begin{figure}
\centering
\begin{subfigure}{0.225\textwidth}
    \includegraphics[width=\textwidth]{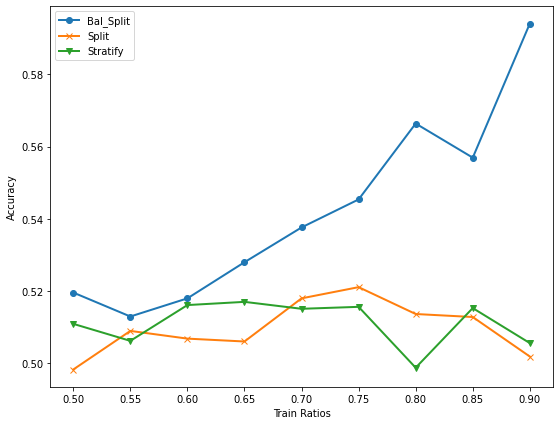}
    \caption{Comparison of accuracy of multiple splitting strategies.}
    \label{Rfacc}
\end{subfigure}
\hfill
\begin{subfigure}{0.225\textwidth}
    \includegraphics[width=\textwidth]{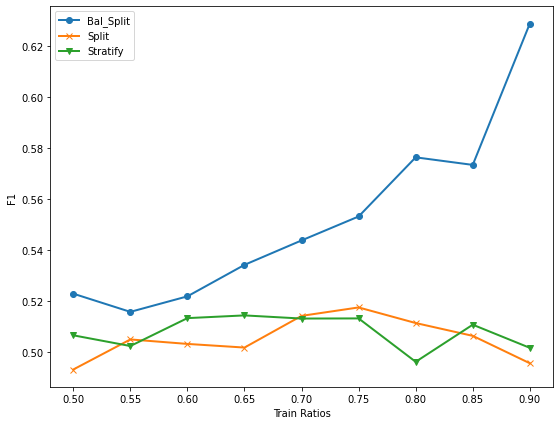}
    \caption{Comparison of F-1 score of multiple splitting strategies.}
    \label{Rff1}
\end{subfigure}
\caption{Comparisons for random forest classifier.}
\label{Rf}
\end{figure}

The results of the random forest classifier's accuracy score and F-1 score are presented in table \ref{tab3}. Figure \ref{Rfacc} compares the random, stratified, and balanced split of the KNN classifier's accuracy score. Figure \ref{Rff1} compares the random, stratified, and balanced split of the KNN classifier's F-1 score.

From the tables and graphs, it can be observed that balanced-split has performed better than random and stratified splits.

\begin{table*}[htbp]
\footnotesize
\centering
\caption{Accuracy score and F-1 score of KNN classifier using random, stratified, and balanced splits.}
\label{tab2}
\begin{tabular}{lllllllllll}
\hline
Split  strategies	&Train ratios &	0.5	&0.55&	0.6	&.65&	0.7&	0.75&	0.8&	0.85&	0.9\\
\hline
Random split&	Accuracy score&	0.412&	0.42	&0.418	&0.419	&0.428	&0.429&	0.431&	0.429	&0.421\\
&	F-1 score&	0.416&	0.429	&0.422	&0.423	&0.431&	0.43&	0.433&	0.431	&0.426\\
\hline
Stratified split	&Accuracy score&	0.4	&0.409&	0.414	&0.419&	0.414	&0.41	&0.416	&0.421	&0.436\\
&	F-1&	0.405&	0.415&	0.419	&0.424&	0.419&	0.415	&0.421	&0.426	&0.437\\
\hline
Balanced split&	Accuracy score&	0.398	&0.411	&0.42	&0.428&	0.43	&0.439&	0.438&	0.436&	0.454\\
&	F-1 score&	0.407	&0.42	&0.43	&0.439&	0.443&	0.454&	0.458&	0.464	&0.501\\
\hline
\end{tabular}
\end{table*}

\begin{table*}[htbp]
\footnotesize
\centering
\caption{Accuracy score and F-1 score of random forest classifier using random, stratified, and balanced splits.}
\label{tab3}
\begin{tabular}{lllllllllll}
\hline
Split stratigies&	train ratios	&0.5	&0.55	&0.6	&0.65&	0.7	&0.75&	0.8&	0.85&	0.9\\
\hline
Random split&	Accuracy score&	0.498&	0.508	&0.506&	0.506	&0.517	&0.521	&0.513&	0.512	&0.501\\
&	F-1 score&	0.493&	0.504&	0.503	&0.501&	0.514&	0.517&	0.511&	0.506&	0.495\\
\hline
Stratified split&	Accuracy score&	0.51&	0.506&	0.516&	0.516&	0.515&	0.515	&0.498	&0.515	&0.505\\
&	F-1 score&	0.506&	0.502&	0.513&	0.514	&0.513&	0.513&	0.496&	0.51&	0.501\\
\hline
Balanced split&	Accuracy score&	0.519	&0.512&	0.517	&0.527&	0.537&	0.545	&0.566	&0.556&	0.594\\
&	F-1 score &0.522	&0.515	&0.521	&0.534	&0.543	&0.553	&0.576	&0.573	&0.628\\
\hline
\end{tabular}
\end{table*}

\section{Conclusion and future works}
We propose a new data-splitting strategy for imbalanced datasets. From the results, it can be concluded that it can be a valuable strategy for imbalanced datasets. Balanced split solves two major existing problems in common splitting strategies. 

We present ideas as potentials for future studies, including further evaluation of balanced-split on image and textual datasets. Comparison of random and stratified-split with data augmentation against balanced-split is is another research idea to explore.

\section*{Code}
We publish the code of balanced-split for further research, and it can be publicly accessed using the following URL: \url{https://github.com/azalahmadkhan/Balanced-split}

\bibliographystyle{IEEEtran}
\bibliography{bibliography.bib} 
\end{document}